\documentclass[11pt,a4paper]{article}
\usepackage[hyperref]{emnlp2020}
\usepackage{times}
\usepackage{latexsym}

\usepackage{balance}

\usepackage{microtype}

\aclfinalcopy % Uncomment this line for the final submission
\newif\ifdebugdoc\debugdoctrue

\ifdebugdoc
\newcommand{\fyi}[1]{\footnote{\textcolor{blue}{fyi:#1}}}

\newcommand{\del}[1]{\textcolor{blue}{\sout{#1}}}
\newcommand{\outline}[1]{\textbf{\colorbox{yellow}{Outline:}\textcolor{red}{#1.}}}

\newcommand{\jie}[1]{\footnote{\colorbox{yellow}{Jie:} #1.}}

\else
\newcommand{\fyi}[1]{}

\newcommand{\del}[1]{}
\newcommand{\jie}[1]{}
\newcommand{\outline}[1]{}
\fi

\newcommand{\nop}[1]{}

\usepackage{xcolor}
\usepackage{balance}
\makeatletter
\g@addto@macro{\UrlBreaks}{\UrlOrds}
\makeatother\usepackage{enumitem}
\usepackage{mathtools}
\usepackage{amssymb}
\usepackage{amsmath}
\usepackage{amsfonts}
\usepackage{booktabs}
\usepackage{array}
\usepackage{multirow}
\usepackage{graphicx}
\newsavebox\CBox
\def\textBF#1{\sbox\CBox{#1}\resizebox{\wd\CBox}{\ht\CBox}{\textbf{#1}}}

\usepackage{soul}
\usepackage{amsthm}
\usepackage[ruled, vlined, linesnumbered]{algorithm2e}

\SetCommentSty{mycommfont}

\DeclareMathOperator*{\argmax}{arg\,max}

\usepackage{makecell}
\usepackage{cases}
\usepackage{scalerel}
\usepackage{hyperref}

\DeclareFixedFont{\ttb}{T1}{txtt}{bx}{n}{12} % for bold
\DeclareFixedFont{\ttm}{T1}{txtt}{m}{n}{12}  % for normal

\usepackage{listings}
\usepackage{xcolor}

\definecolor{codegreen}{rgb}{0,0.6,0}
\definecolor{codegray}{rgb}{0.5,0.5,0.5}
\definecolor{codepurple}{rgb}{0.58,0,0.82}
\definecolor{backcolour}{rgb}{0.95,0.95,0.92}
\definecolor{deepblue}{rgb}{0,0,0.5}
\definecolor{deepred}{rgb}{0.6,0,0}
\definecolor{deepgreen}{rgb}{0,0.5,0}

\usepackage{bm}
\usepackage{adjustbox}

\title{Adversarial Training for Code Retrieval \\with Question-Description Relevance Regularization}

\author{Jie Zhao \\
  The Ohio State University \\
  \texttt{zhao.1359@osu.edu} \\\And
  Huan Sun \\
  The Ohio State University \\
  \texttt{sun.397@osu.edu} \\}

\date{}

\begin{document}
\maketitle
\begin{abstract}
% \hs{other titles to consider:\\ 1. Adversarial Training for Code Retrieval with Question Relevance Regularization\\ 2. Question Relevance Regularized Adversarial Training for Code Retrieval \\ 3. Relevance Regularized Adversarial Training for Code Retrieval\\}
Code retrieval is a key task aiming to match natural and programming languages.
In this work, we propose adversarial learning for code retrieval, that is regularized by question-description relevance.
First, we adapt a simple adversarial learning technique to generate difficult code snippets given the input question, which can help the learning of code retrieval that faces bi-modal and data-scarce challenges.
Second, we propose to leverage question-description relevance to regularize adversarial learning, such that a generated code snippet should contribute more to the code retrieval training loss, only if its paired natural language description is predicted to be less relevant to the user given question.
Experiments on large-scale code retrieval datasets of two programming languages show that our adversarial learning method is able to improve the performance of state-of-the-art models.
Moreover, using an additional duplicate question prediction model to regularize adversarial learning further improves the performance, and this is more effective than using the duplicated questions in strong multi-task learning baselines.\footnote{Source code and dataset are available at https://github.com/jiez-osu/QQC.}
\end{abstract}

% \section{Credits}

% This document has been adapted by Yulan He
% from the instructions for earlier ACL and NAACL proceedings, including those for 
% ACL 2020 by Steven Bethard, Ryan Cotterrell and Rui Yan, 
% ACL 2019 by Douwe Kiela and Ivan Vuli\'{c},
% NAACL 2019 by Stephanie Lukin and Alla Roskovskaya, 
% ACL 2018 by Shay Cohen, Kevin Gimpel, and Wei Lu, 
% NAACL 2018 by Margaret Michell and Stephanie Lukin,
% 2017/2018 (NA)ACL bibtex suggestions from Jason Eisner,
% ACL 2017 by Dan Gildea and Min-Yen Kan, 
% NAACL 2017 by Margaret Mitchell, 
% ACL 2012 by Maggie Li and Michael White, 
% ACL 2010 by Jing-Shing Chang and Philipp Koehn, 
% ACL 2008 by Johanna D. Moore, Simone Teufel, James Allan, and Sadaoki Furui, 
% ACL 2005 by Hwee Tou Ng and Kemal Oflazer, 
% ACL 2002 by Eugene Charniak and Dekang Lin, 
% and earlier ACL and EACL formats written by several people, including
% John Chen, Henry S. Thompson and Donald Walker.
% Additional elements were taken from the formatting instructions of the \emph{International Joint Conference on Artificial Intelligence} and the \emph{Conference on Computer Vision and Pattern Recognition}.

\section{Introduction}
\label{sec:intro}
Recently there has been a growing research interest in the intersection of natural language (NL) and programming language (PL), with exemplar tasks including code generation \cite{Agashe2019JuICeAL,Bi2019IncorporatingEK}, code summarizing \cite{leclair-mcmillan-2019-recommendations, Panthaplackel2020LearningTU}, and code retrieval \cite{gu2018deep}.
In this paper, we study code retrieval, which aims to retrieve code snippets for a given NL question such as \textit{``Flatten a shallow list in Python."}
Advanced code retrieval tools can save programmers tremendous time in various scenarios, such as how to fix a bug, how to implement a function, which API to use, etc. 
Moreover, even if the retrieved code snippets do not perfectly match the NL question, editing them is often much easier than generating a code snippet from scratch. For example, the retrieve-and-edit paradigm \cite{, Hayati2018RetrievalBasedNC, Hashimoto2018ARF, Guo2019CouplingRA} for code generation has attracted growing attention recently, which first employs a code retriever to find the most relevant code snippets for a given question, and then edit them via a code generation model.
Previous work has shown that code retrieval performance can significantly affect the final generated results \cite{Huang2018NaturalLT} in such scenarios.

There have been two groups of work on code retrieval: 
(1) One group of work (e.g., the recent retrieve-and-edit work \cite{Hashimoto2018ARF, Guo2019CouplingRA}) assumes each code snippet is associated with NL descriptions and retrieves code snippets by measuring the relevance between such descriptions and a given question.
(2) The other group of work (e.g., CODENN \cite{iyer2016summarizing} and Deep Code Search \cite{gu2018deep}) directly measures the relevance between a question and a code snippet. 
Comparing with the former group, this group of work has the advantage that they can still apply when NL descriptions are not available for candidate code snippets, as is often the case for many large-scale code repositories \cite{Dinella2020HoppityLG, Chen2019ALS}. 
Our work connects with both groups: We aim to directly match a code snippet with a given question, but during training, we will utilize question-description relevance to improve the learning process.

Despite the existing efforts, we observe two challenges for directly matching code snippets with NL questions, which motivate this work. 
First, code retrieval as a bi-modal task requires representation learning of two heterogeneous but complementary modalities, which has been known to be difficult \cite{Cvitkovic2019OpenVL, leclair-mcmillan-2019-recommendations, Akbar2019SCORSC} and may require more training data. 
This makes code retrieval more challenging compared to document retrieval where the target documents often contain useful shallow NL features like keywords or key phrases.
Second, code retrieval often encounters special one-to-many mapping scenarios, where one NL question can be solved by multiple code solutions that take very different approaches. 
Table~\ref{tab:example} illustrates the challenges.
For $i{=}1{,}2$ or $3$, $q^{\scriptscriptstyle (i)}$ is an NL question{/description} that is associated with a Python answer $c^{\scriptscriptstyle (i)}$. 
Here, question $q^{\scriptscriptstyle (1)}$ should be matched with multiple code snippets: $c^{\scriptscriptstyle (1)}$ and $c^{\scriptscriptstyle (2)}$, because they both flatten a 2D list despite with different programming approaches.
In contrast, $c^{\scriptscriptstyle (3)}$ is performing a totally different task, but uses many overlapped tokens with $c^{\scriptscriptstyle (1)}$.
Hence, it can be difficult to train a code retrieval model that generalizes well to match $q^{\scriptscriptstyle (1)}$ with both $c^{\scriptscriptstyle (1)}$ and $c^{\scriptscriptstyle (2)}$, and is simultaneously able to distinguish $c^{\scriptscriptstyle (1)}$ from $c^{\scriptscriptstyle (3)}$.

To address the first challenge, we propose to introduce adversarial training to code retrieval, which has been successfully applied to transfer learning from one domain to another \cite{tzeng2017adversarial} or learning with scarce supervised data \cite{Kim2019ImageCW}.
Our intuition is that by employing a generative adversarial model to produce \textit{challenging negative code snippets} during training, the code retrieval model will be strengthened to distinguish between positive and negative $\langle q,c\rangle$ pairs.
In particular, we adapt a generative adversarial sampling technique \cite{wang2017irgan}, whose effectiveness has been shown in a wide range of uni-modal text retrieval tasks.

For the second challenge, we propose to further employ \textit{question-description (QD) relevance} as a complementary uni-modal view to reweight the adversarial training samples.
In general, our intuition is that the code retrieval model should put more weights on the adversarial examples that are hard to distinguish by itself, but easy from the view of a QD relevance model.
This design will help solve the one-to-many issue in the second challenge, by differentiating true negative and false negative adversarial examples: If a QD relevance model also suggests that a code snippet is not relevant to the original question, it is more likely to be a true negative, and hence the code retrieval model should put more weights on it.
Note that this QD relevance design aims to help train the code retrieval model better and we do not need NL descriptions to be associated with code snippets at testing phase.

We conduct extensive experiments using a large-scale $\langle$question, code snippet$\rangle$ dataset StaQC \cite{yao2018staqc} and our collected duplicated question dataset from Stack Overflow\footnote{\url{https://stackoverflow.com/}}.
The results show that our proposed learning framework is able to improve the state-of-the-art code retrieval models and outperforms using adversarial learning without QD relevance regularization, as well as strong multi-task learning baselines that also utilize question duplication data.
\begin{table}[t]
    \centering
    % \resizebox{\linewidth}{!}{
    \begin{adjustbox}{max width=\linewidth}
        \begin{tabular}{p{0.15cm}p{7cm}}
            \hline
            \footnotesize \boldsymbol{$q^{\scriptscriptstyle (1)}$} & \footnotesize \textit{Flatten a shallow list in Python} \\
            \footnotesize \boldsymbol{$c^{\scriptscriptstyle (1)}$} & 
\begin{lstlisting}
from itertools import chain
rslt = chain(*list_2d)
\end{lstlisting} \\
            \hline
            \footnotesize \boldsymbol{$q^{\scriptscriptstyle (2)}$} & \footnotesize \textit{How to flatten a 2D list to 1D without using numpy?} \\
            \footnotesize \boldsymbol{$c^{\scriptscriptstyle (2)}$} & 
\begin{lstlisting}
list_of_lists = [[1,2,3],[1,2],[1,4,5,6,7]]
[j for sub in list_of_lists for j in sub]
\end{lstlisting} \\
            \hline
            \footnotesize \boldsymbol{$q^{\scriptscriptstyle (3)}$} & \footnotesize \textit{How to get all possible combinations of a list’s elements?} \\
            \footnotesize \boldsymbol{$c^{\scriptscriptstyle (3)}$} & 
\begin{lstlisting}
from itertools import chain, combinations
subsets = chain(*map(lambda x: combinations(mylist, x), range(0, len(mylist)+1)))
\end{lstlisting} \\
%             \hline
%             \footnotesize \pmb{$q^{\scriptscriptstyle (4)}$} & \footnotesize \textit{Convert multi-dimensional list to a 1D list in Python} \\
%             \footnotesize \pmb{$c^{\scriptscriptstyle (4)}$} & 
% \begin{lstlisting}
% def flatten(listOfLists):
%     return chain.from_iterable(listOfLists)
% \end{lstlisting} \\
            \hline
        \end{tabular}
    % }
    \end{adjustbox}
    % \vspace{-0.2cm}
    \caption{Motivating Example. 
    $\langle q^{\scriptscriptstyle (i)}, c^{\scriptscriptstyle (i)}\rangle$ denotes an associated $\langle$natural language question, code snippet$\rangle$ pair. 
    {$q^{\scriptscriptstyle (i)}$ can also be viewed as a description of $c^{\scriptscriptstyle (i)}$.}
    Given $q^{\scriptscriptstyle (1)}$, the ideal code retrieval result is to return both $c^{\scriptscriptstyle (1)}$ and $c^{\scriptscriptstyle (2)}$ as their programming semantics are equivalent.
    Contrarily, $c^{\scriptscriptstyle (3)}$ is semantically irrelevant to $q^{\scriptscriptstyle (1)}$ and should not be returned, although its surface form is similar to $c^{\scriptscriptstyle (1)}$.
    % However, a code retrieval model can easily make the mistake to find $c^{\scriptscriptstyle (3)}$, which has similar surface form to $c^{\scriptscriptstyle (1)}$, although their logical semantics are irrelevant.
    % In such cases, it can be easier to decide their relationships from the natural language perspective, because $q^{\scriptscriptstyle (1)}$ is more similar to $q^{\scriptscriptstyle (2)}$ than $q^{\scriptscriptstyle (3)}$.
    In such cases, it can be easier to decide their relationships from the question perspective, because $\langle q^{\scriptscriptstyle (1)}$, $q^{\scriptscriptstyle (2)}\rangle$ are more alike than $\langle q^{\scriptscriptstyle (1)}$, $q^{\scriptscriptstyle (3)}\rangle$.
    }
    \label{tab:example}
    % \vspace{-0.2cm}
\end{table}

\section{Overview}
\label{sec:problem-def}
The work studies \textit{\textbf{code retrieval}}, a task of matching questions with code, which we will use \textbf{QC} to stand for.
The training set $\mathcal{D^\text{\scriptsize{QC}}}$ consists of NL question and code snippet pairs $\mathcal{D^\text{\scriptsize{QC}}} {=} \{q^{\scriptscriptstyle (i)}, c^{\scriptscriptstyle (i)}\}$. 
Given NL question $q^{\scriptscriptstyle (i)}$, the QC task is to find $c^{(\scriptscriptstyle i)}$ from $\mathcal{D}^\text{\scriptsize{QC}}$ among all the code snippets.
For simplicity, we omit the data sample index and use $q$ and $c$ to denote a QC pair, and $c^-$ to represent any other code snippets in the dataset except for $c$.

Our goal is to learn a QC model, denoted as $f_\theta^\text{\scriptsize{QC}}$, that retrieves the highest score code snippets for an input question:
$\argmax_{c' \in \{c\} \cup \{c^-\}} {f_\theta^\text{\scriptsize{QC}}(q, c')}$.
Note that at testing time, the trained QC model $f^\text{\scriptsize{QC}}$ can be used to retrieve code snippets from any code bases, unlike the group of QC methods \cite{Hayati2018RetrievalBasedNC,Hashimoto2018ARF,Guo2019CouplingRA} relying on the availability of NL descriptions of code.

We aim to address the aforementioned challenges in code retrieval through two strategies: 
(1) We introduce adversarial learning \cite{goodfellow2014generative} to alleviate the bi-modal learning challenges. Specifically an adversarial QC generator selects unpaired code snippets that are difficult for the QC model to discriminate, to strengthen its ability to distinguish top-ranked positive and negative samples \cite{wang2017irgan}. 
(2) We also propose to employ a question-description (\textbf{QD}) relevance model to provide a secondary view on the generated adversarial samples, inspired by the group of QC work that measures the relevance of code snippets through their associated NL descriptions.

\begin{figure}[t]
    \centering
    \resizebox{\columnwidth}{!}{
        \includegraphics[width=\columnwidth]{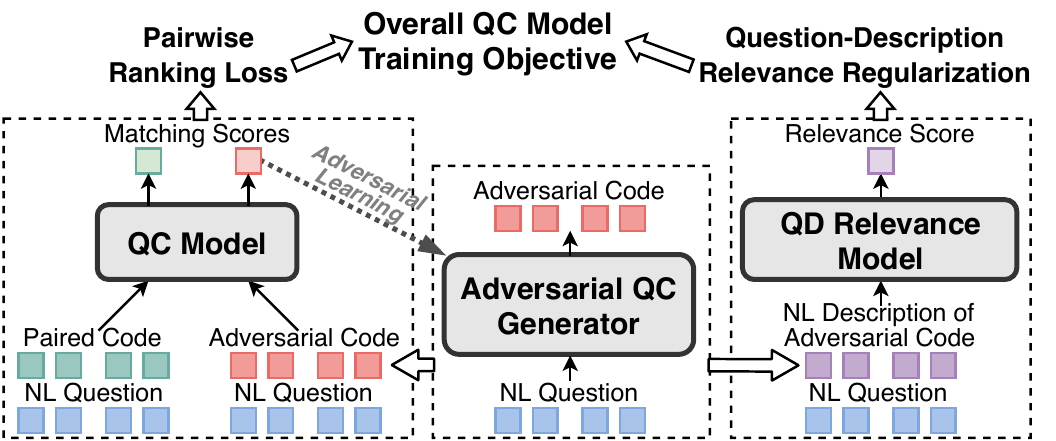}
    }
    % \vspace{-0.5cm}
    \caption{
        Regularized adversarial learning framework. Best viewed in color. 
        The adversarial QC generator (middle) produces an adversarial code given an NL question.
        The QD relevance model (right) then predicts a relevance score between the given question and the NL description or the generated adversarial code.
        A pairwise ranking loss is computed between the ground-truth code and the adversarial code.
        The QC model (left) is trained with the ranking loss, after it is scaled by a QD relevance regularization weight that depends on the QD relevance score.
        The parameter update is larger when the relevance score is smaller and vice versa.
    }
    \label{fig:framework}
    % \vspace{-0.4cm}
\end{figure}

Figure~\ref{fig:framework} gives an overview of our proposed learning framework, which does not assume specific model architectures and can be generalized to different base QC models or use different QD relevance models.
A general description is given in the caption.
In summary, the adversarial QC generator selects $\hat{c}$ that is unpaired with a given $q$.
$\hat{q}$ is an NL description of $\hat{c}$. Details on how to acquire $\hat{q}$ will be introduced in Section~\ref{sec:algo-relevance}.
Next, a QD model predicts a relevance score for $\langle q, \hat{q}\rangle$.
A pairwise ranking loss is calculated based on whether the QC model discriminates ground-truth QC pair $\langle q, c\rangle$ from unpaired $\langle q, \hat{c}\rangle$.
Learning through this loss is reweighted by a down-scale factor, which is dynamically determined by the QD relevance prediction score.
This works as a regularization term over potential false negative adversarial samples.

\section{Proposed Methodology}
We now introduce in detail our proposed learning framework. We start with the adversarial learning method in Section~\ref{sec:algo-advsample} and then discuss the rationale to incorporate question-description or QD relevance feedback in Section~\ref{sec:algo-relevance}, before putting them together in Section~\ref{sec:algo} and Section~\ref{sec:basemodel}.

\subsection{Adversarial Learning via Sampling}
\label{sec:algo-advsample}
We propose to apply adversarial learning \cite{goodfellow2014generative} to code retrieval.
Our goal is to train a better QC model ${f_\theta^\text{\scriptsize{QC}}}$ by letting it play the adversarial game with a QC generator model ${g_\phi^\text{\scriptsize{QC}}}$.
$\theta$ represents the parameters of the QC model and $\phi$ represents the parameters of the adversarial QC generator.
As in standard adversarial learning, ${f_\theta^\text{\scriptsize{QC}}}$ plays the discriminator role to distinguish ground-truth code snippet $c$ from generated pairs $\hat{c}$. 
The training objective of the QC model is to minimize $\mathcal{L_\theta}$ below:
\setlength{\abovedisplayskip}{5pt}%
\setlength{\belowdisplayskip}{5pt}%
\setlength{\abovedisplayshortskip}{5pt}%
\setlength{\belowdisplayshortskip}{5pt}%
\setlength{\jot}{0pt}% Inter-equation spacing
\begin{align*}
    \mathcal{L}_\theta &= \sum_i \mathbb{E}_{\hat{c} \sim {P_\phi(c|q^{\scriptscriptstyle (i)})}} l_\theta(q^{\scriptscriptstyle (i)}, c^{\scriptscriptstyle (i)}, \hat{c}) ,\\
    l_\theta &= \text{max}(0, d{+}f_\theta^\text{\scriptsize{QC}}(q^{\scriptscriptstyle (i)},\hat{c}){-}f_\theta^\text{\scriptsize{QC}}(q^{\scriptscriptstyle (i)},c^{\scriptscriptstyle (i)})) ,
\end{align*}
where $l_\theta$ is a pairwise ranking loss, and specifically we use a hinge loss with margin $d$.
$\hat{c}$ is generated by ${g_\phi^\text{\scriptsize{QC}}}$ and follows a probability distribution $P_\phi(c|q^{\scriptscriptstyle (i)})$. 
${g_\phi^\text{\scriptsize{QC}}}$ aims to assign higher probabilities to code snippets that would mislead ${f_\theta^\text{\scriptsize{QC}}}$.

There are many ways to realize the QC generator. For example, one may employ a sequence model to generate the adversarial code snippet $\hat{c}$ token by token \cite{Bi2019IncorporatingEK, Agashe2019JuICeAL}.
However, training a sequence generation model is difficult, because the search space of all code token combinations is huge.
Henceforce, we turn to a simpler idea inspired by \citet{wang2017irgan}, and restrict the generation of $\hat{c}$ to the space of all the existing code snippets in the training dataset $\mathcal{D}^\text{\scriptsize{QC}}$.
The QC generator then only needs to sample an existing code snippet $c^{\scriptscriptstyle (j)}$ from an adversarial probability distribution conditioned on a given query and let it be $\hat{c}$, i.e., $\hat{c}{=}c^{\scriptscriptstyle (j)}{\sim}P_\phi(c|q^{\scriptscriptstyle (i)})$.
Adopting this method will make training the QC generator easier, and ensures that the generated code snippets are legitimate as they directly come from the training dataset.
We define the adversarial code distribution as:
\setlength{\abovedisplayskip}{5pt}%
\setlength{\belowdisplayskip}{5pt}%
\setlength{\abovedisplayshortskip}{0pt}%
\setlength{\belowdisplayshortskip}{0pt}%
\begin{align*}
    P_\phi(c|q^{\scriptscriptstyle (i)}) &= \frac{\text{exp}({g_\phi^\text{QC}}(q^{\scriptscriptstyle (i)}, c) / \tau)}{\sum_{c'} \text{exp}({g_\phi^\text{QC}}(q^{\scriptscriptstyle (i)}, c') / \tau)} \label{eq:p_c_given_q} \ ,
\end{align*}
where ${g_\phi^\text{QC}}$ represents an adversarial QC matching function.
$\tau$ is a temperature hyper-parameter used to tune the distribution to concentrate more of less on top-scored code snippets. 
Moreover, scoring all code snippets can be computationally inefficient in practice. Therefore, we use the method of \citet{yang2019adversarial} to first uniformly sample a subset of data, whose size is much smaller than the entire training set size, and then perform adversarial sampling on this subset.

The generator function $g_\phi^\text{\scriptsize{QC}}$ can be pre-trained in the same way as the discriminator (i.e., $f_\theta^\text{\scriptsize{QC}}$) and then get updated using standard policy gradient reinforcement learning algorithms, such as REINFORCE \cite{williams1992simple}, to maximize the ranking losses of the QC model.
Formally, the QC generator aims to maximize the following expected reward:
$J(\phi) {=} \sum_{i} \mathbb{E}_{c^{\scriptscriptstyle (j)} \sim P_\phi(c|q^{\scriptscriptstyle (i)})} [l_\theta(q^{\scriptscriptstyle (i)}, c^{\scriptscriptstyle (i)}, c^{\scriptscriptstyle (j)})]$,
where $l_\theta(q^{\scriptscriptstyle (i)}, c^{\scriptscriptstyle (i)}, c^{\scriptscriptstyle (j)})$ is the pairwise ranking loss of the discriminator model defined earlier.
The gradient of $J$ can be derived as $\nabla_\phi J {=} \sum_{i} \mathbb{E}_{c^{\scriptscriptstyle (j)} \sim P_\phi(c|q^{\scriptscriptstyle (i)})} [ l_\theta \cdot \nabla_\phi \ log P_\phi(c^{\scriptscriptstyle (j)}|q^{\scriptscriptstyle (i)})]$.
Another option is to let $g_\phi^\text{\scriptsize{QC}}$ use the same architecture as $f_\theta^\text{\scriptsize{QC}}$ and use tied parameters (i.e., $\phi{=}\theta$), as adopted in previous work \cite{deshpande2019dissecting, park2019adversarial}.

The focus of this work is to show the effectiveness of applying adversarial learning to code retrieval, and how to regularize it with QD relevance.
We leave more complex adversarial techniques (e.g. adversarial perturbation \cite{goodfellow2014explaining, miyato2015distributional} or adversarial sequence generation \cite{Li2018GeneratingDA}) for future studies.

\subsection{Question-Description Relevance Regularization}
\label{sec:algo-relevance}

% Because of the one-to-many mapping issue of code retrieval, some adversarially generated samples can be false negative and may hurt the learning of a QC model. 
% Intuitively, we can train a better code retrieval model, if the negative code snippets are all true-negative ones: code snippets that are confusingly similar to correct code answers, but performs different functionalities.
Intuitively, we can train a better code retrieval model, if the negative code snippets are all true-negative ones, i.e., if they are confusingly similar to correct code answers, but perform different functionalities.
However, because of the one-to-many mapping issue, some negative code snippets sampled by the adversarial QC generator can be false-negative, i.e. they are equally good answers for a given question despite that they are not paired with the question in the training set. 
Unfortunately during training, this problem could become increasingly obvious as the adversarial will be improved along with the code retrieval model, and eventually makes learning less and less effective.
Since both the QC model and the adversarial QC generator operates from the QC perspective, it is difficult to further discriminate true-negative and false-negative code snippets.

Therefore, we propose to alleviate this problem with QD relevance regularization.
This idea is inspired by the group of QC work mentioned in Section~\ref{sec:intro} that retrieves code snippets by matching their NL descriptions with a given question.
But different from them, we only leverage QD relevance during training to provide a secondary view and to reweight the adversarial samples. 
Fortunately, an adversarial code snippet $\hat{c}$ sampled from the original training dataset $\mathcal{D}^\text{\scriptsize{QC}}$ is paired with an NL question $\hat{q}$, which can be regarded as its NL description and used to calculate the relevance to the given question $q$.

Let us refer to the example in Table~\ref{tab:example} again. 
At a certain point of training, with $q^{\scriptscriptstyle (1)}$ ``\textit{Flatten a shallow list in Python}" being the given question, the adversarial QC generator may choose $c^{\scriptscriptstyle (2)}$ and $c^{\scriptscriptstyle (3)}$ as the negative samples, but {instead of treating them equivalently, we can} infer from the QD matching perspective that $c^{\scriptscriptstyle (3)}$ is likely to be true negative, because  $q^{\scriptscriptstyle (3)}$ ``\textit{How to get all possible combinations of a list’s elements}" clearly has different meanings from $q^{\scriptscriptstyle (1)}$, while $c^{\scriptscriptstyle (2)}$ is likely to be a false negative example since $q^{\scriptscriptstyle (2)}$ ``\textit{How to flatten a 2D list to 1D without using numpy?}" is similar to $q^{\scriptscriptstyle (1)}$. 
Hence, during training, the discriminative QC model should put more weights on negative samples like $c^{\scriptscriptstyle (3)}$ rather than $c^{\scriptscriptstyle (2)}$.

We now explain how to map QD relevance scores to regularization weights.
Let $f^\text{\scriptsize{QD}}(q, \hat{q})$ denote the predicted relevance score between the given question $q$ and the question paired with an adversarial code snippet $\hat{q}$, and let $f^\text{\scriptsize{QD}}(q, \hat{q})$ be normalized to the range from 0 to 1.
We can see from the above example that QD relevance and adjusted learning weight should be reversely associated, so we map the normalized relevance score to a weight using a monotonously decreasing polynomial function: $w^\text{\scriptsize QD}(x){=}(1 {-} x^a)^b,\ 0{\leq}x{\leq}1$.
Both $a$ and $b$ are positive integer hyper-parameters that control the shape of the curve and can be tuned on the dev sets. 
In this work, they are both set to one by default for simplicity. 
$w^\text{\scriptsize QD} \in [0,1]$ allows the optimization objective to weigh less on adversarial samples that are more likely to be false negative. 

\begin{algorithm}[t]
\small
\SetAlgoLined
\SetKwInOut{InputQC}{QC training data}
\SetKwInOut{InputQD}{QD model}
\SetKwInOut{InputConst}{Constants}
    \InputQC{$\mathcal{D}^\text{\tiny{QC}} = \{q^{\scriptscriptstyle (i)}, c^{\scriptscriptstyle (i)}\}$}
    \InputQD{$f^\text{\tiny QD}$}
    \InputConst{positive intergers $N$, $\tau$, $a$, $b$}
    \KwResult{QC model $f_\theta^\text{\tiny{QC}}$}
    $\triangleright$ Pretrain $f_\theta^\text{\tiny{QC}}$ on $\mathcal{D}^\text{\tiny{QC}}$ using pairwise ranking loss $l_\theta^\text{QC}$ with randomly sampled negative code snippets \;
    $\triangleright$ Initialize QC generator ${g_\phi^\text{\tiny{QC}}}$ with $f_\theta^\text{\tiny{QC}}$: $\phi \leftarrow \theta$ \;
    \While{not converge or not reach max iter number}{
        \For{random sampled $\langle q^{\scriptscriptstyle (i)}, c^{\scriptscriptstyle (i)} \rangle \in \mathcal{D}^\text{\tiny{QC}}$}{
            Randomly choose $D {=} \{q{,}c\} \subset \mathcal{D}^\text{\tiny{QC}}$, where $|D|{=}N$\;
            Sample $c^{\scriptscriptstyle (j)} {\in} D$, that $c^{\scriptscriptstyle (j)} \sim P_\phi(c^{\scriptscriptstyle (j)}|q^{\scriptscriptstyle (i)}) = \text{softmax}_\tau({g_\phi^\text{\tiny QC}}(q^{\scriptscriptstyle (i)}, c^{\scriptscriptstyle (j)}))$ \;
            $l_\theta^\text{QC} \leftarrow l_\theta(q^{\scriptscriptstyle(i)}, c^{\scriptscriptstyle(i)}, c^{\scriptscriptstyle(j)})$ \;
            Find $q^{\scriptscriptstyle (j)}$ associated with $q^{\scriptscriptstyle (j)}$, $w^\text{\tiny QD} \leftarrow (1 - f^\text{\tiny QD}(q^{\scriptscriptstyle (i)}, q^{\scriptscriptstyle (j)})^a)^b$ \;
            Update QC model with gradient descent to reduce loss: $w^\text{\tiny QD} \cdot l_\theta^\text{\tiny QC}$ \;
            Update adversarial QC generator with gradient ascent: $l^\text{\tiny QC}_\theta \cdot \nabla_\phi \ log P_\phi(c^{\scriptscriptstyle (j)}|q^{\scriptscriptstyle (i)})$ \
        }
        $\triangleright$ {Optional QD model update. (See Section~\ref{sec:basemodel}.)}
    }
\caption{Question-Description Relevance Regularized Adversarial Learning.}
\label{algo:main}
\end{algorithm}

\subsection{Question-Description Relevance Regularized Adversarial Learning}
\label{sec:algo}
Now we describe the proposed learning framework in Algorithm~\ref{algo:main} that combines adversarial learning and QD relevance regularization.
Let us first assume the QD model is given and we will explain how to pre-train, and optionally update it shortly.

The QC model can be first pre-trained on $\mathcal{D}^\text{\scriptsize{QC}}$ using standard pairwise ranking loss $l_\theta(q^{\scriptscriptstyle (i)}, c^{\scriptscriptstyle (i)}, c^{\scriptscriptstyle (j)})$ with randomly sampled $c^{\scriptscriptstyle (j)}$.
Line 3-11 show the QC model training steps.
For each QC pair $\langle q^{\scriptscriptstyle (i)}, c^{\scriptscriptstyle (i)}\rangle$, a batch of negative QC pairs are sampled randomly  from the training set $\mathcal{D}^\text{\scriptsize{QC}}$. 
The QC generator then choose an adversarial $c^{\scriptscriptstyle (j)}$ from distribution $P_\phi(c|q^{\scriptscriptstyle (i)})$ defined in Section~\ref{sec:algo-advsample}, and its paired question is $q^{\scriptscriptstyle (j)}$.
Two questions $q^{\scriptscriptstyle (i)}$ and $q^{\scriptscriptstyle (j)}$ are then passed to the QD model, and the QD relevance prediction is mapped to a regularization weight $w^\text{\scriptsize QD}$.
Finally, the regularization weight is used to control the update of the QC model on the ranking loss with the adversarial $\hat{c}$.

\subsection{Base Model Architecture}
\label{sec:basemodel}
Our framework can be instantiated with various model architectures for QC or QD.
Here we choose the same neural network architecture as \cite{gu2018deep, yao2019coacor} as our base QC model, that achieves competitive or state-of-the-art code retrieval performances.
Concretely, both a natural language question $q$ and a code snippet $c$ are sequences of tokens. They are encoded respectively by separate bi-LSTM networks \cite{schuster1997bidirectional}, passed through a max pooling layer to extract the most salient features of the entire sequence, and then through a hyperbolic tangent activate function.
The encoded question and code representations are denoted as $h^q$ and $h^c$.
Finally, a matching component scores the vector representation between $q$ and $c$ and outputs their matching score for ranking. We follow previous work to use cosine similarity: $f^\text{\scriptsize{QC}}(q,c) = \text{cosine}(h^q, h^c)$.

\noindent\textbf{QD Model.}
There are various model architecture choices, but here for simplicity, we adapt the QC model for QD relevance prediction.
We let the QD model use the same neural architecture as the QC model, but with Siamese question encoders.
The QD relevance score is the cosine similarity between $h^{q^{(i)}}$ and $h^{q^{(j)}}$, the bi-LSTM encoding outputs for question $q^{\scriptscriptstyle (i)}$ and $q^{\scriptscriptstyle (j)}$ respectively: $f^\text{\scriptsize{QD}}(q^{\scriptscriptstyle (i)}{,}q^{\scriptscriptstyle (j)}) {=} \text{cosine}(h^{q^{\scriptscriptstyle (i)}}{,}h^{q^{\scriptscriptstyle (j)}})$. 
This method allows using a pre-trained QC model to initialize the QD model parameters, which is easy to implement and the pre-trained question encoder in the QC model can help the QD performance.
Since programming-domain question paraphrases are rare, we collect a small QD training set consisting of programming related natural language question pairs $\mathcal{D^\text{\scriptsize{QD}}} {=} \{q^{\scriptscriptstyle (j)}, p^{\scriptscriptstyle (j)}\}$ based on duplicated questions in Stack Overflow. 
% See Section \ref{sec:exp-dataset} for details.

The learning framework can be symmetrically applied, as indicated by Line 12 in Algorithm~\ref{algo:main}, so that the QD model can also be improved.
This may provide better QD relevance feedback to help train a better QC model.
In short, we can use a discriminative and a generative QD model. The generative QD model selects adversarial questions to help train the discriminative QD model, and this training can be regularized by the relevance predictions from a QC model. 
More details will be introduced in the experiments.
\section{Experiments}
In this section, we first introduce our experimental setup, and then will show that our method not only outperforms the baseline methods, but also multi-task learning approaches, where question-description relevance prediction is the other task.
In particular, the QD relevance regularization consistently improves QC performance upon adversarial learning, and the effectiveness of relevance regularization can also be verified as it is symmetrically applied to improve the QD task.

\subsection{Datasets}
\label{sec:exp-dataset}
We use StaQC \cite{yao2018staqc} to train and evaluate our code retrieval model, which contains automatically extracted\nop{detected ``how-to-do-it" type of} questions on Python and SQL and their associated code answers from Stack Overflow.
We use the version of StaQC that each question is associated with a single answer, as those associated with multiple answers are predicted by an automatic answer detection model and therefore noisier. 
We randomly split this QC datasets by a 70/15/15 ratio into training, dev and testing sets.
The dataset statistics are summarized in Table~\ref{tab:datastats}.

% Note that there exist code retrieval datasets between $<$method/function description, method/function implementation$>$ pairs, such as \cite{gu2016deep, gu2018deep}. These datasets are automatically curated with limited filtering based on simple heuristics. As found in previous work \cite{cambronero2019deep,husain2019codesearchnet}, method descriptions are fundamentally different from code-related queries, because they are composed by the same code authors and may not accurately describe the associated code. In addition, retrieving full methods or functions rather than arbitrary code snippets is only a simplified approximation of the general code search task.

We use Stack Exchange Data Explorer\footnote{\href{https://data.stackexchange.com/stackoverflow/query/new}{SEDE} 
and \href{https://meta.stackexchange.com/questions/2677/database-schema-documentation-for-the-public-data-dump-and-sede}{SEDE schema documentation.}}
to collect data for training and evaluating QD relevance prediction.
Specifically, we collect the question pairs from posts that are manually labeled as duplicate by users, which are related by \texttt{LinkTypeId=3}.
It turns out that the QD datasets are substantially smaller than the QC datasets, especially for Python, as shown in Table~\ref{tab:datastats}. 
This makes it more interesting to check whether a small amount of QD relevance guidance can help improve code retrieval performances.

\begin{table}[t]
    \centering
    \resizebox{0.9\columnwidth}{!}{
        \begin{tabular}{l|rrr|rrr}
            \toprule
            & \multicolumn{3}{c|}{Python} & \multicolumn{3}{c}{SQL} \\
            & \textbf{Train} & \textbf{Dev} & \textbf{Test} & \textbf{Train} & \textbf{Dev} & \textbf{Test} \\ 
            \hline
            \textbf{QC} & 68,235 & 8,529 & 8,530 & 60,509 & 7,564 & 7,564 \\  
            \textbf{QD} & 1,085 & 1,085 & 1,447 & 18,020 & 2,252 & 2,253 \\
            \bottomrule
        \end{tabular}
    }
    % \vspace{-0.3cm}
    \caption{Dataset statistics. QD is used to represent the duplicate question dataset.}
    \label{tab:datastats}
\end{table}

\subsection{Baselines and Evaluation Metrics}
We select state-of-the-art methods from both groups of work for QC (mentioned in Introduction).
DecAtt and DCS below are methods that directly match questions with code. 
EditDist and vMF-VAE transfer code retrieval into a question matching problem.
\begin{itemize}[leftmargin=*, noitemsep, topsep=0pt]
    \item DecAtt \cite{parikh-etal-2016-decomposable}. This is a widely used neural network model with attention mechanism for sentence pairwise modeling.
    \item DCS \cite{gu2018deep}. We use this as our base model, because it is a simple yet effective code retrieval model that achieves competitive performance without introducing additional training overheads \cite{yao2019coacor}. Its architecture has been described in Section~\ref{sec:basemodel}.
    \item EditDist \cite{Hayati2018RetrievalBasedNC}. Code snippets are retrieved by measuring an edit distance based similarity function between their associated NL descriptions and the input questions. 
    Since there is only one question for each sample in the QC datasets, we apply a standard code summarization tool \cite{iyer2016summarizing} to generate code descriptions to match with input questions.
    \item vMF-VAE \cite{Guo2019CouplingRA}. This is similar to EditDist, but a vMF Variational Autoencoder \cite{Xu2018Spherical} is separately trained to embed questions and code descriptions into latent vector distributions, whose distance is then measured by KL-divergence. This method is also used by \citet{Hashimoto2018ARF}.
\end{itemize}
We further consider multi-task learning (MTL) as an alternative way how QD can help QC.
It is worth mentioning that our method does not require \nop{sharing of \emph{labeled}}{\emph{associated}} training data or the sharing of trained parameters between the QD and QC tasks, whereas MTL {typically} does.  
For fair comparison, we adapt two MTL methods to our scenario that use the same base model, or its question and code encoders:
\begin{itemize}[leftmargin=*, noitemsep, topsep=0pt]
    \item MTL-DCS. This is a straightfoward MTL adaptation of DCS, where the code encoder is updated on the QC dataset and the question encoder is updated on both QC and QD datasets. The model is alternatively trained on both datasets.
    \item MTL-MLP \cite{gonzalez2018strong}. This recent MTL method is originally designed to rank relevant questions and question-related comments. It uses a multi-layer perceptron (MLP) network with one shared hidden layer, a task-specific hidden layer and a task-specific classification layer for each output. We adapt it for our task.
    The input to the MLP is the concatenation of similarity features $[max(h^q, h^c), h^q - h^c, h^q \odot h^c]$, where $\odot$ is element-wise product. $h^q$ and $h^c$ are learned using the same encoders as our base model.
\end{itemize}
    
The ranking metrics used for evaluation are Mean Average Precision (MAP) and Normalize Discounted Cumulative Gain (nDCG) \cite{jarvelin2002cumulated}.
The same evaluation method as previous work is adopted \cite{iyer2016summarizing, yao2019coacor} for both QC and QD, where we randomly choose from the testing set a fixed-size (49) pool of negative candidates for each question, and evaluate the ranking of its paired code snippet or questions among these negative candidates.

\subsection{Implementation Details}
Our implementation is based on \citet{yao2019coacor}.
We follow this work to set the base model hyper-parameters.
The vocabulary embedding size for both natural language and programming language is set at 200.
The LSTM hidden size is 400.
Margin in the hinge loss is 0.05.
The trained DCS model is used as pre-training for our models.
The learning rate is set at 1e-4 and the dropout rate set at 0.25.
For adversarial training, we set $\tau$ to 0.2 following \cite{wang2017irgan} and limit the maximum number of epochs to 300.
Standard L2-regularization is used on all the models.
We empirically tried to tie the parameters of the discriminator and the generator following previous work \cite{deshpande2019dissecting, park2019adversarial}, which shows similar improvements over the baselines.
Implementation from \citet{Xu2018Spherical} is used for the vMF-VAE baseline.

We follow the code preprocessing steps in \citet{yao2018staqc} for Python and \citet{iyer2016summarizing} for SQL. 
We use the NLTK toolkit \cite{bird-loper-2004-nltk} to tokenize the collected duplicate questions, and let it share the same NL vocabulary as the QC dataset $\mathcal{D}^\text{\scriptsize QC}$.

% We use Stack Exchange Data Explorer \footnote{\url{https://data.stackexchange.com/stackoverflow/query/new}} to collect data for training and testing question relevance prediction.
% Specifically, we collect the question pairs from posts that are manually labeled as duplicate by users, i.e., which are related by \texttt{LinkTypeId=3}.
% More details about Stack Exchange Data Exploreo can be found here: \url{https://meta.stackexchange.com/questions/2677/database-schema-documentation-for-the-public-data-dump-and-sede/2678#2678}.
% More implementation details can be found in the supplementary material.

\begin{table}[t]
    \centering
    \resizebox{\columnwidth}{!}{
    \begin{tabular}{l|cc|cc}
        \toprule
        & \multicolumn{2}{c|}{Python} & \multicolumn{2}{c}{SQL} \\
        & \textbf{MAP} & \textbf{nDCG} & \textbf{MAP} & \textbf{nDCG} \\ 
        \hline
        \textbf{EditDist} \cite{Hayati2018RetrievalBasedNC} & 0.2348 & 0.3844 & 0.2096 & 0.3641 \\
        \textbf{vMF-VAE} \cite{Guo2019CouplingRA} & 0.2886 & 0.4511 & 0.2921 & 0.4537 \\
        \textbf{DecAtt} \cite{parikh-etal-2016-decomposable} & 0.5744 & 0.6716 & 0.5142 & 0.6231 \\ 
        % \textbf{CODE-NN} \cite{iyer2016summarizing} & -- & -- & 0.5004 & 0.6096 \\  
        \textbf{DCS} \cite{gu2018deep} & 0.6015 & 0.6929 & 0.5155 & 0.6237 \\  
        % \hline
        \hline
        \textbf{MTL-MLP} \cite{gonzalez2018strong} & 0.5737 & 0.6712 & 0.5079 & 0.6179 \\
        \textbf{MTL-DCS} & 0.6024 & 0.6935 & 0.5160 & 0.6237 \\
        \hline
        \textbf{Our} & \textBF{0.6372}$^*$ & \textBF{0.7206}$^*$ & \textBF{0.5404}$^*$ & \textBF{0.6429}$^*$ \\
        \textbf{Our} - RR & 0.6249$^*$ & 0.7111$^*$ & 0.5274$^*$ & 0.6327$^*$ \\
        % \textbf{Our -- Fix QQ} & 0.6359 & 0.7197 & 0.5362 & 0.6396 \\
        % \textbf{Our -- Fix QQ} & 0.6371 & 0.7205 & 0.5366 & 0.6398 \\
        \bottomrule
    \end{tabular}
    }
    % \vspace{-0.3cm}
    \caption{Code retrieval (QC) performance on test sets. * denotes significantly different from DCS \cite{gu2018deep} in one-tailed t-test ($p < 0.01$).}
    % \vspace{-0.1cm}
    \label{tab:qc}
\end{table}

\begin{figure}
    \centering
    % \vspace{-0.32cm}
    \includegraphics[width=\columnwidth]{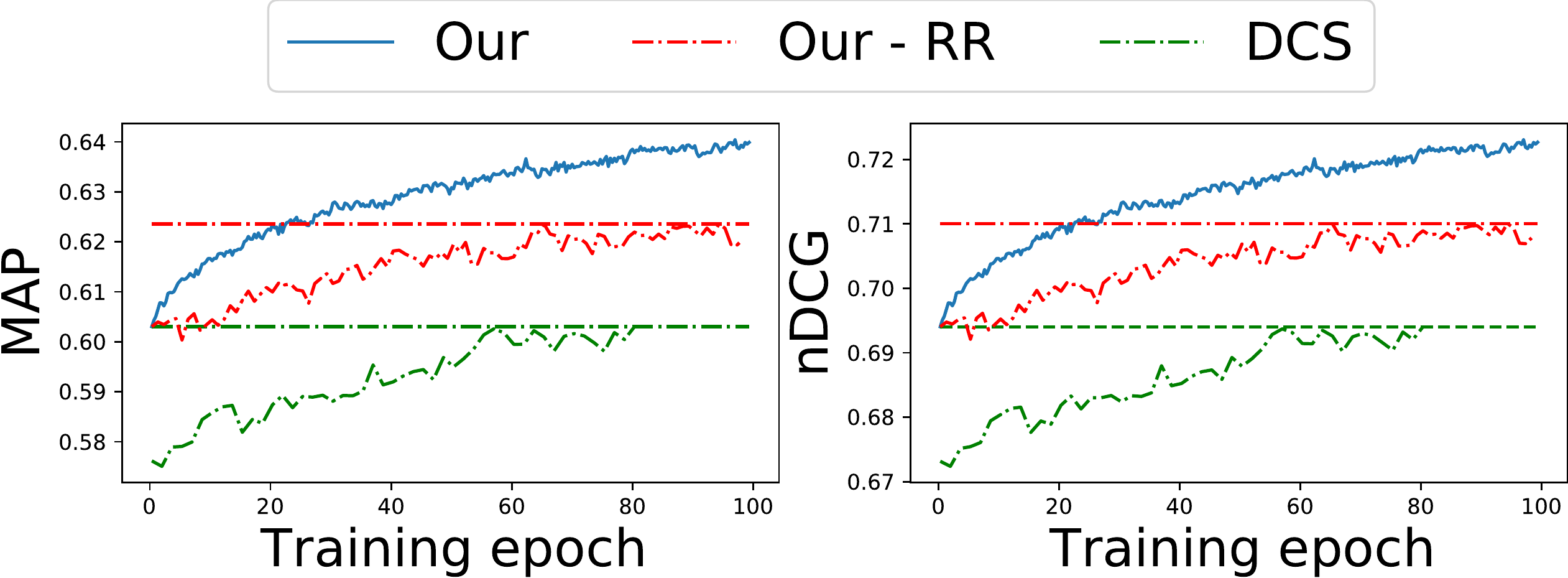}
    % \vspace{-0.8cm}
    \caption{QC learning curves on the Python dev set.}
    \label{fig:training}
\end{figure}

\subsection{Results and Analyses}
Our experiments aim to answer the following research questions:

\noindent (1) \textit{Can the question regularized adversarial learning framework improve code retrieval (QC) performance?}
We will first compare the code retrieval performance of different methods. 
Table~\ref{tab:qc} summarizes the \textit{test results}, which are consistent on both Python and SQL datasets.
Code retrieval baselines by measuring QD relevance, e.g., EditDist and vMF-VAE, are popularly used in code generation related work, but do not perform well compared to other code retrieval baselines in our experiments, partly because they are not optimized toward the QC task. 
This suggests that applying more advanced code retrieval methods for retrieve-and-edit code generation can be an interesting future research topic.
DCS is a strong baseline, as it outperforms DecAtt that uses a more complex attention mechanism. 
This indicates that it is not easy to automatically learn pairwise token associations between natural language and programming languages from software community data, which is also suggested by previous work \cite{Panthaplackel2019AssociatingNL, vinayakarao2017anne}.

Our proposed learning algorithm can improve the QC performance compared to all the baselines.
The ``- RR" variant is to only apply adversarial sampling without QD relevance regularization.
It already leads to improvements compared to the base model (i.e. DCS), but does not perform as well as our full model.
This proves the usefulness of the QD relevance regularization and indicates that selectively weighting the contribution of adversarial samples to the training loss can help the model generalize better to test data.
Figure~\ref{fig:training} compares QC learning curves on the \textit{dev set}.
The full model curve being the smoothest qualitatively suggests that the adversarial learning has been well regularized.

\noindent (2) \textit{How does the proposed algorithm compare with multi-task learning methods?}
The results are reported in Table~\ref{tab:mtl}.
The MTL-MLP model is originally proposed to improve question-question relevance prediction by using question-comment relevance prediction as a secondary task \cite{gonzalez2018strong}. It does not perform as well as MTL-DCS, which basically uses hard parameter sharing between the two tasks and does not require additional similarity feature definitions.
In general, the effectiveness of these MTL baselines on the QC task is limited because there are only a small amount of QD pairs available for training. 
Both our method and its ablated variant outperform the MTL baselines. 
This shows that it may be more effective to use a data scarce task to regularize the adversarial learning of a relatively data rich task, than using those scarce data in MTL.
\begin{table}[t]
    \centering
    \resizebox{\columnwidth}{!}{
    \begin{tabular}{l|cc|cc}
        \toprule
        & \multicolumn{2}{c|}{Python} & \multicolumn{2}{c}{SQL} \\
        & \textbf{MAP} & \textbf{nDCG} & \textbf{MAP} & \textbf{nDCG} \\ 
        \hline
        % \textbf{EditDist} \cite{Hayati2018RetrievalBasedNC} & 0.2348 & 0.3844 & 0.2096 & 0.3641 \\
        % \textbf{VAE} \cite{Guo2019CouplingRA} & 0.2886 & 0.4511 & 0.2921 & 0.4537 \\
        % \hline
        % \textbf{DecAtt} \cite{parikh-etal-2016-decomposable} & 0.5744 & 0.6716 & 0.5142 & 0.6231 \\ 
        % % \textbf{CODE-NN} \cite{iyer2016summarizing} & -- & -- & 0.5004 & 0.6096 \\  
        % \textbf{DCS} \cite{gu2018deep} & 0.6015 & 0.6929 & 0.5155 & 0.6237 \\  
        % \hline
        \textbf{MTL-MLP} \cite{gonzalez2018strong} & 0.5737 & 0.6712 & 0.5079 & 0.6179 \\
        \textbf{MTL-DCS} & 0.6024 & 0.6935 & 0.5160 & 0.6237 \\
        \hline
        % \textbf{Our - QR} & 0.6249 & 0.7111 & 0.5274 & 0.6327 \\
        % \textbf{Our -- Fix QQ} & 0.6359 & 0.7197 & 0.5362 & 0.6396 \\
        % \textbf{Our -- Fix QQ} & 0.6371 & 0.7205 & 0.5366 & 0.6398 \\
        \textbf{Our} & \textBF{0.6372} & \textBF{0.7206} & \textBF{0.5404} & \textBF{0.6429} \\
        \bottomrule
    \end{tabular}
    }
    % \vspace{-0.3cm}
    \caption{Compare QC performance with MTL.}
    % \vspace{-0.2cm}
    \label{tab:mtl}
\end{table}

\noindent (3) \textit{Can the QD performance be improved by the proposed method?}
Although QD is not the focus of this work, we can use it to verify that generalizability of our method by symmetrically applying it to update the QD model as mentioned in Section~\ref{sec:algo-relevance}.
To be concrete, a generative adversarial QD model selects difficult questions from the a distribution of question pair scores: $\hat{q} \sim \text{softmax}_\tau(f^\text{\scriptsize QD}(\hat{q},q^{\scriptscriptstyle (i)}))$.
Then a QC model is used to calculate a relevance score for a question-code pair, and this can regularize the adversarial learning of the QD model.

Table~\ref{tab:qq} shows the results. 
Our method and its ablated variants outperform the QD baselines EditDist and vMF-VAE, again suggesting that supervised learning is more effective.
The full model achieves the best overall performance and removing relevance regularization (- RR) from the QC model consistently leads to performance drop.
In contrast, further removing adversarial sampling (- AS) hurts the performance on SQL dataset slightly, but not on Python.
This is probably because the Python QD dataset is very small and using adversarial learning can easily overfit, which again suggests the importance of our proposed relevance regularization.
Finally, removing QC as pretraining (- Pretrain) greatly hurts the performance, which is understandable since QC datasets are much larger.

Because the QD model performance can be improved in such a way, we allow it to get updated in our QC experiments (corresponding to line 12 in Algorithm~\ref{algo:main}) and the results have been discussed in Table~\ref{tab:qc}.
We report here the QC performance using a fixed QD model (i.e. Our - RR - AS) for relevance regularization: MAP=0.6371, nDCG=0.7205 for Python and MAP=0.5366, nDCG=0.6398 for SQL.
Comparing these results with those in Table\ref{tab:qc} (Our), one can see that allowing the QD model to update consistently improves QC performance, which suggests that a better QD model can provide more accurate relevance regularization to the QC model and leads to better results.

\begin{table}[t]
    \centering
    \resizebox{\columnwidth}{!}{
        \begin{tabular}{l|cc|cc}
            \toprule
            & \multicolumn{2}{c|}{Python} & \multicolumn{2}{c}{SQL} \\
            & \textbf{MAP} & \textbf{nDCG} & \textbf{MAP} & \textbf{nDCG} \\ 
            \hline
            \textbf{EditDist} \cite{Hayati2018RetrievalBasedNC} & 0.3617 & 0.4883 & 0.3246 & 0.4580 \\
            \textbf{vMF-VAE} \cite{Guo2019CouplingRA} & 0.3009 & 0.4616 & 0.3029 & 0.4641 \\
            \hline
            \textbf{Our} & \textBF{0.7162} & \textBF{0.7821} & \textBF{0.6947} & \textBF{0.7651} \\
            \textbf{Our} - RR & 0.7046 & 0.7734 & 0.6846 & 0.7575 \\
            \textbf{Our} - RR - AS  & 0.7116 & 0.7787 & 0.6764 & 0.7512 \\  
            \textbf{Our} - RR - AS - Pretrain & 0.3882 & 0.5170 & 0.6284 & 0.7129 \\  
            % \textbf{Our} & 0.4278 & 0.5488 & 0.6562 & 0.7344 \\
            % \hline\hline
            % \textbf{QC-DCS} & 0.6770 & 0.6736 & 0.7484 & 0.5578 & 0.5538 & 0.6493 \\  
            % \hline
            % \textbf{MTL-MLP} \cite{gonzalez2018strong} & 0.6811 & 0.7470 & \textBF{0.6949} & \textBF{0.7655} \\
            % \textbf{MTL-DCS} & 0.6767 & 0.7510 & 0.5637 & 0.6591 \\
            % \hline
            \bottomrule
        \end{tabular}
    }
    % \vspace{-0.3cm}
    \caption{Question relevance prediction results, evaluated on the question duplication dataset we collected.}
    \label{tab:qq}
    % \vspace{-0.2cm}
\end{table}

\section{Related Work}
\label{sec:related}

\noindent\textbf{Code Retrieval.}
Code retrieval has developed from using classic information retrieval techniques \cite{hill2014nl, haiduc2013automatic, lu2015query} to recently deep neural methods that can be categorized into two groups.
The first group directly model the similarity across the natural language and programming language modalities.
Besides CODENN \cite{iyer2016summarizing} and DCS \cite{gu2018deep} discussed earlier, \citet{yao2019coacor} leverage an extra code summarization task and ensemble a separately trained code summary retrieval model with a QC model to achieve better overall code retrieval performances. \citet{ye2020leveraging} further train a code generation model and a code summarization model through dual learning, which helped to learn better NL question and code representations. 
Both works employ additional sequence generation models that greatly increases the training complexity, and they both treat all unpaired code equally as negatives.
Our work differs from them as we introduce adversarial learning for code retrieval, and the existing work do not leverage question relevance for code retrieval as we do.

The second group of works transfer code retrieve to a code description retrieval problem, similar to general domain question answering, where two natural language sentences are matched \cite{zhao2017end, zhao2019riker}.
This methodology has been widely adopted as a component in the retrieve-and-edit code generation literature.
For example, heuristic methods such as measuring edit distance \cite{Hayati2018RetrievalBasedNC} or comparing code type and length \cite{Huang2018NaturalLT} are used, and separate question latent representations \cite{Hayati2018RetrievalBasedNC, Guo2019CouplingRA} are learned. 
Our work shares with them the idea to exploit QD relevance, but we use QD relevance in a novel way to regularize the adversarial learning of QC models.
It will be an interesting future work to leverage the proposed code retrieval method for retrieve-and-edit code generation.

\noindent\textbf{Adversarial Learning.}
Adversarial learning has been widely used in areas such as computer vision \cite{Mirza2014ConditionalGA, Chen2016InfoGANIR, Radford2015UnsupervisedRL, Arjovsky2017WassersteinGA}, text generation \cite{yu2017seqgan, chen2019bofgan, liang2019unsupervised, gu2018deep, Liu2017AdversarialML, ma2019detect, zhao2019easy}, relation extraction \cite{Wu2017AdversarialTF, Qin2018DSGANGA}, question answering \cite{Oh2019OpenDomainWA, yang2019adversarial}, etc.
We proposed to apply adversarial learning to code retrieval, because they have effectively improved cross-domain task performances and helped generate useful training data, 
We adapted the method from \citet{wang2017irgan} for the bi-modal QC scenario.
As future work, adversarial learning for QC can be generalized to other settings with different base neural models \cite{yang2019adversarial} or with more complex adversarial learning methods, such as adding perturbed noises \cite{park2019adversarial} or generating adversarial sequences \cite{yu2017seqgan, Li2018GeneratingDA}.
Our method differs from most adversarial learning work in that the discriminator (QC model) does not see all generated samples as equally negative.
\section{Conclusion}
This work studies the code retrieval problem, and tries to tackle the challenges of matching natural language questions with programming language (code) snippets.
We propose a novel learning algorithm that introduces adversarial learning to code retrieval, and it is further regularized from the perspective of a question-description relevance prediction model. 
Empirical results show that the proposed method can significantly improve the code retrieval performances on large-scale datasets for both Python and SQL programming languages.

\section*{Acknowledgments}
We would like to thank the anonymous reviewers for their helpful comments. This research was sponsored in part by the Army Research Office under cooperative agreements W911NF-17-1-0412, NSF Grant IIS1815674, and NSF CAREER \#1942980. The views and conclusions contained herein are those of the authors and should not be interpreted as representing the official policies, either expressed or implied, of the Army Research Office or the U.S. Government. The U.S. Government is authorized to reproduce and distribute reprints for Government purposes notwithstanding any copyright notice herein.

% \newpage
\balance
\bibliographystyle{acl_natbib}
\bibliography{main}

\balance
% \newpage
% \input{tex/supplementary}
\end{document}